\documentclass[sigconf]{acmart}

\usepackage{algorithm}
\usepackage{algorithmic}
\usepackage{multirow}
\usepackage{amsmath}
\usepackage{xspace}
\usepackage{microtype}
\usepackage{comment}

\usepackage{enumitem}

\newcommand{\modelname}{LargeGNN\xspace}
\newcommand{\datasetname}{DBpedia1M\xspace}

\newtheorem{definition}{Definition}

\DeclareMathOperator*{\argmax}{argmax}

\AtBeginDocument{%
  \providecommand\BibTeX{{%
    \normalfont B\kern-0.5em{\scshape i\kern-0.25em b}\kern-0.8em\TeX}}}

\copyrightyear{2022} 
\acmYear{2022} 
\setcopyright{acmcopyright}\acmConference[CIKM '22]{Proceedings of the 31st ACM International Conference on Information and Knowledge Management}{October 17--21, 2022}{Atlanta, GA, USA}
\acmBooktitle{Proceedings of the 31st ACM International Conference on Information and Knowledge Management (CIKM '22), October 17--21, 2022, Atlanta, GA, USA}
\acmPrice{15.00}
\acmDOI{10.1145/3511808.3557374}
\acmISBN{978-1-4503-9236-5/22/10}

\begin{document}

\title{Large-scale Entity Alignment via Knowledge Graph Merging, Partitioning and Embedding}

\author{Kexuan Xin}
\affiliation{
    \institution{The University of Queensland \country{Australia}}
}
\email{uqkxin@uq.edu.au}

\author{Zequn Sun}
\authornote{Equal contributors}
\affiliation{
  \institution{Nanjing University \country{China}}
}
\email{zqsun.nju@gmail.com}

\author{Wen Hua}
\authornotemark[1]
\authornote{Corresponding author}
\affiliation{%
  \institution{The University of Queensland \country{Australia}}
}
\email{w.hua@uq.edu.au}

\author{Wei Hu}
\affiliation{
  \institution{Nanjing University \country{China}}
}
\email{whu@nju.edu.cn}

\author{Jianfeng Qu}
\affiliation{
  \institution{Soochow University \country{China}}
}
\email{jfqu@suda.edu.cn}

\author{Xiaofang Zhou}
\affiliation{
  \institution{Hong Kong University of Science and Technology \country{HKSAR}}
  }
\email{zxf@cse.ust.hk}

\renewcommand{\shortauthors}{Kexuan Xin et al.}

\begin{abstract}
Entity alignment is a crucial task in knowledge graph fusion.
However, most entity alignment approaches have the scalability problem.
Recent methods address this issue by dividing large KGs into small blocks for embedding and alignment learning in each. 
However, such a partitioning and learning process results in an excessive loss of structure and alignment.
Therefore, in this work, we propose a scalable GNN-based entity alignment approach to reduce the structure and alignment loss from three perspectives.
First, we propose a centrality-based subgraph generation algorithm to recall some landmark entities serving as the bridges between different subgraphs.
Second, we introduce self-supervised entity reconstruction to recover entity representations from incomplete neighborhood subgraphs, and design cross-subgraph negative sampling to incorporate entities from other subgraphs in alignment learning. 
Third, during the inference process, we merge the embeddings of subgraphs to make a single space for alignment search.
Experimental results on the benchmark OpenEA dataset and the proposed large DBpedia1M dataset verify the effectiveness of our approach.
\end{abstract}



\maketitle
\section{Introduction}
Knowledge graphs (KGs) describe the structured facts of entities in a multi-relational structure and have been successfully applied to a variety of AI-driven applications \cite{KG_survey, qu2021noise}. 
Due to its incompleteness, a single KG usually cannot satisfy the knowledge requirement.
Integrating multiple KGs draws more attention because they usually share common facts and cover complementary information.
An effective way to integrate heterogeneous KGs is through entity alignment \cite{suchanek2011paris}. 
It seeks to find identical entities in different KGs that refer to the same real-world object. 
Recently, embedding-based entity alignment approaches have achieved great progress \cite{sun2017cross,sun2018bootstrapping,cao2019multi,wu2019relation,sun2020knowledge,zeng2020degree,mao2020relational,xin2022informed, xin2022ensemble}.
One major advantage of embedding techniques is that they eliminate the need for hand-crafted features or rules.
KG embeddings have also shown great strength in dealing with the symbolic heterogeneity of different KGs \cite{wang2017knowledge}.
Especially for entity alignment, embedding-based approaches seek to capture the similarity of entities in vector space \cite{zhao2020experimental,EA_survey_AIOpen}. 

Among existing entity alignment studies, graph neural networks (GNNs) have demonstrated their superiority of encoding graph structures and capturing entity similarities, 
and achieved promising performance \cite{yu2021knowledge, wu2019relation, sun2020knowledge, mao2020relational, mao2021boosting}. 
The main reason of the good performance of GNN-based entity alignment is the superiority of GNNs in capturing the isomorphic structures between KGs.
Despite the significant progress made by existing GNN-based EA approaches, they all suffer from scalability issues.
The difficulty of implementing GNN-based entity alignment approaches grows with the scale of KGs, owing to the exponential growth of node neighboring size.
For example, we found that RREA \cite{mao2020relational}, KEGCN \cite{yu2021knowledge} and AliNet \cite{sun2020knowledge} all fail on the OpenEA \cite{sun2020benchmarking} 100K datasets with 11GB-memory GPU due to their large memory requirements for graph convolution over the whole KG. 
However, 100K is far from the size of real-life KGs, which may contain tens of millions of entities.
The scalability issue severely limits the applicability of GNN-based entity alignment approaches.

To address the scalability issue, recent studies \cite{ge2021largeea, zeng2022entity} have attempted to mitigate the memory usage problem by partitioning the two KGs into several small blocks and performing embedding learning, alignment learning, and alignment search within each block.
LargeEA \cite{ge2021largeea} first divides the source and target KGs to be aligned into several subgraphs independently.
Then, it pairs the source and target subgraphs to construct entity alignment blocks by measuring the recall of the seed entity alignment pairs.
LIME \cite{zeng2022entity} utilizes a bidirectional partition method to protect KG structures.
It doubles the partition time of LargeEA while still having the same issue of potentially generating a garbage block that has no potential entity alignment. 
Furthermore, LargeEA and LIME both use entity names to aid in blocking or alignment learning, which has been shown to be not robust due to the name bias issue \cite{liu2020exploring}.
We argue that such ``blocking-then-learning'' framework has two major limitations.
First, dividing a graph into subgraphs and performing embedding learning on the subgraph level results in the \textbf{KG structure loss} because the discarded triples during partitioning would never be used.
Second, the framework is heavily reliant on the partition quality, and a poor partition would cause the \textbf{entity alignment loss}.
Once the partition is completed, the matched entities in different blocks will never be aligned.

Towards effective and efficient large-scale entity alignment, we elaborately design a scalable approach that trains the GNN with subgraphs as mini-batches and infers entity alignment with the approximate $k$NN search across the entire entity embedding space.
The main idea is that the message passing process in GNNs does not have to go through the whole graph, because only some of the entities are important for generating the representation of an entity.
We merge the two KGs into a large graph by combining the pre-aligned entities in seed alignment into a new node.
It can guarantee all the pre-aligned entities to be positioned in the same subgraph and simplify the training process among subgraphs.
We then partition the joint graph into several subgraphs as the mini-batches to train the GNN for reducing the memory cost.
We compensate for the structure and alignment loss from three perspectives:
\begin{list}{\labelitemi}{\leftmargin=1em}
    \item First, we design a centrality-based subgraph generation algorithm to recall some landmark entities in each subgraph.
    Hence, our subgraphs are not disjoint with each other.
    The landmark entities can recover some of the lost KG structures and alignment pairs.
    The duplicates of landmark entities also allow messages to be sent across subgraphs so that the GNN can capture global structures.
    \item Second, considering that we cannot completely avoid the structure loss in KG partition (i.e., an entity may lose some edges in a subgraph),
    we design a self-supervised entity reconstruction objective to generate entity representations from its incomplete neighborhood or its duplicate's neighborhood in another subgraph.
    Besides, different from existing work that limits the alignment learning in a subgraph, 
    we propose cross-subgraph negative sampling to incorporate entities from other subgraphs in alignment learning. 
    Hence, our embedding and alignment learning objectives have both local and global views.
    \item Third, in the inference process, we merge and fuse the embeddings of subgraphs to build a unified entity embedding space,
    so that the lost entity alignment pairs in the KG partition are likely to be found in the space.
    Furthermore, considering that the real entity alignment scenario has many non-matchable entities, we propose a bidirectional approximate $k$NN search in the large space to select high-confidence entity alignment pairs as output.
\end{list}
We construct a new million-scale entity alignment dataset, DBpedia1M, based on DBpedia \cite{DBpedia}. 
It also consists of a large amount of non-matchable entities, which is closer to the real-world KG distribution and has higher difficulty for large-scale entity alignment.
We evaluate the superiority of our approach on both DBpedia1M and the benchmark OpenEA dataset \cite{sun2020benchmarking}. 
Experimental results demonstrate the superiority of our approach that improves the scalability of entity alignment in both space (memory usage) and time.

\section{Related Work}
\subsubsection*{\textbf{Embedding-based Entity Alignment}}
With the growth of deep learning techniques, embedding-based entity alignment approaches have achieved successful progress.
One main direction of embedding-based entity alignment approaches is structure-based, which solely utilizes the structure of KGs. 
There are three types of structure-based approaches. 
Early approaches focus on making use of relational structure learning, which is based on TransE \cite{bordes2013translating}, such as MTransE \cite{chen2017multilingual}, AlignE \cite{sun2018bootstrapping}, SEA \cite{pei2019semi}. 
Some approaches explore the long-term relational dependency of entities, such as IPTransE \cite{zhu2017iterative}, RSN4EA \cite{guo2018recurrent}, IMEA \cite{xin2022informed}. 
With the great achievement of GNN, recent studies focus more on GNN-based embedding techniques, such as GCN-Align \cite{wang2018cross}, AliNet \cite{sun2020knowledge}, RREA \cite{mao2020relational} and Dual-AMN \cite{mao2021boosting}, and achieve the state-of-the-art performance. 
This is because the neighborhood integration can effectively aggregate useful information, which has the superiority to deal with graph structures.
However, GNN-based approaches suffer from poor scalability when dealing with large-scale graphs.
Therefore, 
we elaborately design a GNN-based approach for entity alignment in large-scale KGs.

Besides, some approaches incorporate additional information to enhance alignment learning, such as attribute triples \cite{trisedya2019entity, sun2017cross}, entity names \cite{wu2019relation, liu2020exploring}, images \cite{liu2021visual}, descriptions \cite{chen2018co, tang2020bertint, yang2019aligning, zhang2019multi} and distant supervision information from pre-trained language models \cite{yang2019aligning, tang2020bertint}. 
However, the limited availability of side information and poor robustness restrict the generalizability of these approaches. 
By contrast, our approach primarily considers structure information.

\subsubsection*{\textbf{Large-scale Entity Alignment}}
The scalability issue is a long-standing and tricky problem for the entity alignment task \cite{sun2020benchmarking, zhao2020experimental}.
Although some light-weight approaches like MTransE \cite{chen2017multilingual} and GCN-Align \cite{wang2018cross} are fast and memory-saving, they have poor performance due to over-simple modeling structures.
On the contrary, GNN-based models are more effective but suffer from the scalability issue.
There are some GNN studies handling single large-scale graph, such as GraphSAGE \cite{hamilton2017inductive}, ClusterGCN \cite{chiang2019cluster}, GraphSAINT \cite{zeng2020degree}, etc.
However, these studies are not suitable for multiple KGs. 
LargeEA \cite{ge2021largeea} is the first attempt to conduct entity alignment on large-scale KGs.
It significantly reduces memory usage while relying on entity name to compensate for structure loss caused by partitioning.
LIME \cite{zeng2022entity} is another recent research on scalable entity alignment. 
It improves the partition of LargeEA, and further improves the efficiency of the alignment inference stage.
However, both of them conduct entity alignment on subgraph level and totally lose the global information. 
In this work, we are devoted to developing a GNN specifically for large-scale entity alignment and complementing the lost global structure information without using entity names.

\begin{figure*}\centering
\includegraphics[width=0.9\linewidth]{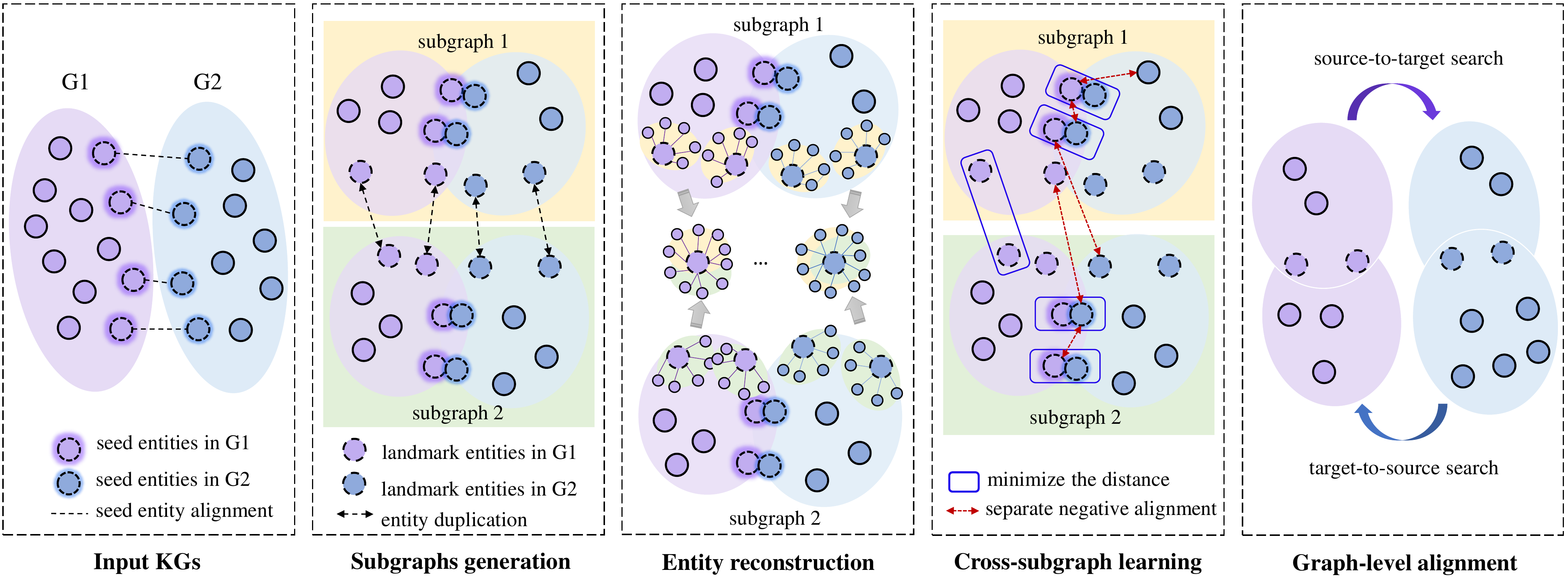}
\caption{\label{fig:framework}
Pipeline of the proposed scalable GNN-based approach for entity alignment in large KGs.}
\end{figure*}

\section{Approach Overview}
Let a KG be represented as $\mathcal{G}=(\mathcal{E}, \mathcal{R}, \mathcal{T})$, 
which is a directed multi-relational graph with a set of entities $\mathcal{E}$ and relations $\mathcal{R}$, as well as a set of triples $\mathcal{T} = \{(h,r,t)\ |\ h,t \in \mathcal{E}, \ \text{and}\ r \in \mathcal{R}\}$. 
Given a source KG $\mathcal{G}_1$ and a target KG $\mathcal{G}_2$, 
the goal of entity alignment is to find as many aligned entity pairs as possible, i.e., $\mathcal{A} = \{(e_1, e_2) \in \mathcal{E}_1 \times \mathcal{E}_2| e_1 \equiv e_2\}$, 
where $\equiv$ indicates equivalence.
In most cases, a small ratio of pre-aligned entity pairs, i.e., seed entity alignment $\mathcal{A}_s \subset \mathcal{A}$, is given in advance as training data.

The pipeline of our approach is shown in Figure~\ref{fig:framework}.
We first merge two KGs into a single graph, and then divide it into multiple small subgraphs using the proposed centrality-based subgraph generation algorithm.
Each subgraph includes entities and triples from the source and target KGs.
We also recall some landmark entities in subgraphs that act as bridges to allow messages to pass between subgraphs.
The GNN learning stage has two goals: entity reconstruction from incomplete neighborhood subgraphs and alignment learning with negative sampling across subgraphs to allow global information propagation across the entire KG structures.
Finally, we combine the entity embeddings of subgraphs into a single embedding space and perform graph-level alignment inference to increase the coverage of potential mappings.

\section{Centrality-based Subgraph Generation}\label{sect:subgraph_gen}
\begin{figure*}
    \centering
    \includegraphics[width=1.0\linewidth]{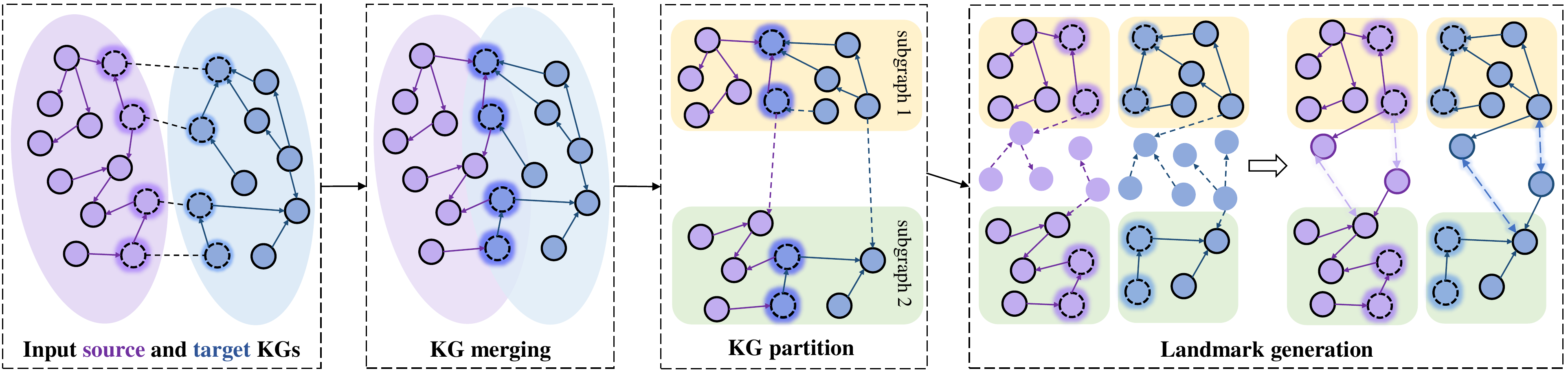}
    \caption{\label{fig:g_partition}Illustration of the proposed subgraph generation procedure.}
\end{figure*}

To make large KGs trainable, the first step is to partition them into several subgraphs so that each subgraph can be fit into the memory of the existing machine.
The complete subgraph generation process is shown in Figure~\ref{fig:g_partition}.
We first convert two graphs into a single graph by merging the seed entity pairs.
Then, we divide the single graph into several subgraphs.
Next, to complement the structure loss brought by partitioning, we recall some landmark entities in each subgraph.
We generate the candidate entities (denoted as $C$) to be recalled from the neighboring entities within two hops of the subgraph.
We design an entity recalling algorithm to select landmark nodes from $C$, to get the final subgraphs.

\begin{algorithm}[t]
\caption{Centrality-based subgraph generation}
\label{alg:entrality-based}
\begin{flushleft}
  \textbf{Input}: Two KGs $\mathcal{G}_1$ and $\mathcal{G}_2$, as well as the training data $\mathcal{A}_s$ and the number of subgraphs $n$.\\
  \textbf{Output}: Subgraph set $\mathbf{G}$.\\
\end{flushleft}
\begin{algorithmic}[1] 
\STATE Conduct KG partition based on $\mathcal{A}_\text{train}$\ to acquire a set of initial subgraphs $\mathbf{G}^0 = \{S_1, \dots, S_n\}$;
\STATE Annotate the compatibility score of each entity via Eq. (\ref{equ: importance});
\STATE Acquire the influential score of each node via Eq. (\ref{equ: infl});
\STATE $\mathbf{G} \leftarrow \emptyset$;
\FOR{$i=1 \rightarrow n$}
  \STATE $C_i \leftarrow$ Acquire the candidate entity set of $S_i$;
  \STATE $L_i \leftarrow$ Select landmark entities from $C_i$;
  \STATE $S_i \leftarrow S_i \cup L_i$; \xspace $\mathbf{G} \leftarrow \mathbf{G} \cup \{S_i\}$;

\ENDFOR
\STATE \textbf{return} $\mathbf{G}$;
\end{algorithmic}
\end{algorithm}
\subsection{Seed-targeted KG Merging and Partitioning}
The primary challenge of partitioning a large graph comes from how to minimize the structure loss of the resulting subgraphs.
A widely-used graph partition strategy is METIS \cite{karypis1997metis}, which aims to decompose the graph into several subgraphs with vertices evenly distributed while minimizing the number of discarded edges (those that are across subgraphs).
However, for the entity alignment task, there are two graphs involved, which are connected by entity alignment pairs. 
As the essence of the alignment process is to bring the pre-aligned entities closer together, we should divide the two graphs into small subgraphs with entities from both the source and target KGs in each.
Given a small set of entity alignment pairs as training data in advance, 
and the fact that equivalent entities tend to have more similar neighboring entities, 
we propose a simple but effective KG merging and partitioning method for entity alignment.
We convert the two KGs into a single graph by merging each pair of seed entity alignment, and then partition this joint graph using the METIS algorithm.
In this way, we can guarantee that all pre-aligned entities in the training set are located in the same subgraph. 
As all of the training entity pairs are preserved, and the aligned entity pairs have more aligned neighboring entities, the remaining test pairs can also be mostly preserved in the subgraph. This simple strategy is quite effective based on our preliminary experiments. 
It largely reduces the computation complexity compared to the existing solutions (LargeEA \cite{ge2021largeea} and LIME \cite{zeng2022entity}). 

\subsection{Centrality-based Entity Benefit}
Even though METIS can effectively decrease the edge cut number, there is still some structure loss inevitably being incorporated.
To make up for the loss of structure in subgraphs, we carefully design the important entity recalling algorithm to find landmark entities that can be added to multiple subgraphs.
The shared landmark entities across subgraphs can serve as the bridging medium during the training strategy.
The recalling procedure is illustrated in algorithm \ref{alg:entrality-based}.
Firstly, we define the importance of each entity after acquiring the initial partitions of the two KGs.
Then, we annotate the influential score of each entity.
Finally, we introduce the centrality-based entity recalling algorithm to select the most beneficial landmark entities and recall them back to each subgraph to complement the structure loss.
In the following, we introduce how to define the benefit of recalling an entity in a given subgraph.

\subsubsection{Importance-based entity compatibility}
\label{sec: importance}
Intuitively, each entity has a different importance with respect to the graph structure information.
In our scenario, seed entities carry more information for alignment learning, and the information is propagated from seed entities to their neighboring entities hop by hop.
Since information spreads from central entities to their neighbors and the rate of loss is high, we define the importance of an entity as:
\begin{equation}
    \phi(e) = \frac{1}{\eta + \min_{e' \in \mathcal{A}_s}d(e, e')}
    \label{equ: importance}
\end{equation}
where $d(e, e')$ denotes the number of hops away from $e$ to $e'$,
and $\eta$ denotes a smoothing factor.
Based on this formula, seed entities have the largest importance score, and for others, the closer they are to seed entities, the higher their importance.

\subsubsection{Centrality-based influential entities}
Inspired by Eigenvector centrality \cite{negre2018eigenvector} that the node is more influential if it has more connections to important nodes, we define the influential score of an entity as the sum of the importance of its neighboring entities:
\begin{equation}
\label{equ: infl}
    \Phi(e) = \sum_{e' \in N_e}\phi(e'),
\end{equation}
where $N_e$ denote the set of neighbors of $e$.
By this definition, an entity is more influential if it is connected with more entities that contain alignment information (e.g., seed entities).

\subsubsection{Benefit of recalling an entity}
Next, we start the landmark recalling procedure for each subgraph.
The standard for recalling an entity is defined as the amount of benefit that recalling this entity can bring to the given subgraph.
We define the benefit of an entity $e$ with respect to a subgraph $S$ as follows:
\begin{equation}
\label{benefit}
    \Omega(e, S) = \Phi(e) \times \lambda^{\mathcal{D}(e, S)}
\end{equation}
where $\mathcal{D}(e, S)$ is the minimum number of hops between the entity $e$ and $\forall \ e'\in S$. 
$\lambda \in (0,1)$ is a decay factor.
Intuitively, if an entity has a high influence score and it is close to the subgraph, it would be more beneficial to recall this entity and add it to the subgraph.

\subsection{Connectivity-aware Landmark Generation}
Given the benefit score of each entity, we then start the landmark entity generation procedure.
Since checking all entities outside a given subgraph is too time-consuming, we first generate an entity candidate set for the given subgraph. 
We consider all the two-hop neighbors of the subgraph as the candidates.
Other entities far away from the subgraph have few useful information, but bring exponential computation complexity. We formally define the problem of landmark selection before going into the algorithm details.

\begin{definition}[Candidate landmark set]
The set of candidate landmark entities $C$ is the union of two-hop neighboring entities of the given subgraph, i.e., $C=\{e\,|\,e \notin S, \mathcal{D}(e, S)\le 2\}$.
\end{definition}

\begin{definition}[Top-$k$ Landmark Search]
Given a subgraph $S$ and a budget $k$, we aim to select $k$ most beneficial entities from $C$ to form the final landmark set $L$, where $L$ needs to guarantee that:
\begin{equation}
    \label{eq:landmark_search}
    L^* = \argmax_{L} \sum_{e \in L} \Omega(e, S), 
    \ s.t. \ 
    \begin{cases}
        |L| \le k\\
        \forall \ e \in L, e \leadsto S
    \end{cases}
\end{equation}
where $e \leadsto S$ means that entity $e$ is connected with subgraph $S$. 
The goal is to select $k$ landmarks from $C$, where the total benefit $\sum_{e \in L}\Omega(e, S)$ is maximized and there are no isolated entities in $L$.
\end{definition}

\subsubsection{Entity recalling procedure.} Straightforwardly, we can enumerate all the ${|C|}\choose{k}$ possible combinations of candidate entities and find $L^*$ with the largest overall benefit. However, this brute-force solution is too time-consuming. We introduce an efficient method for landmark entity generation, as shown in Algorithm \ref{alg:recalling}. We start by ranking all entities in $C$ by descending order of their benefit score $e.\Omega$ (lines 3-4). It is obvious that we can easily select the top-$k$ candidates from this ranked list as landmarks if they are all 1-hop neighbors of the subgraph $S$. However, the main challenge lies in how to guarantee the connectivity of 2-hop candidates. In order to add an potentially isolated (2-hop) candidate $e$ into $L$, its 1-hop neighbor $e'$ must be included in $L$ already so that $e$ can connect to $S$ through $e'$ (In particular, we only need to check its 1-hop neighbor with the highest benefit score $e'=e.max_{nei}$). Otherwise, we hold the entity pair $(e,e')$ in an intermediate dictionary $I$ for later consideration (lines 8-17). Intuitively, such entity pair can be added into $L$ only if its average benefit $avg(e.\Omega,e'.\Omega)$ is larger than the currently top-ranked 1-hop candidate in $C$ (lines 18-24).

\begin{algorithm}[t]
\caption{Connectivity-aware landmark entity generation.}
\label{alg:recalling}
\begin{flushleft}
  \textbf{Input}: A subgraph $\mathcal{S}$ and its candidate landmark set $C$.\\
  \textbf{Parameter}: The budget $k$.\\
  \textbf{Output}: Landmark entity set $L$.\\
\end{flushleft}
\begin{algorithmic}[1] 
\STATE $I \leftarrow \emptyset$; // \textit{Intermediate dictionary}
\STATE $inter \leftarrow 0$; // \textit{Max benefit score of entity pairs in $I$}
\STATE Calculate $\Omega(e, S)$ for each $e \in C$ based on Eq. (\ref{benefit});
\STATE Rank entities in $C$ in decreasing order of $\Omega(e, S)$;
\STATE $L \leftarrow \emptyset$; $i=1$;
\WHILE{$i < |C| \ \& \ |L| < k$}
  \STATE $e = C[i]$;
\IF{$e.hop > 1$} 
  \STATE \textit{/* Deal with potentially isolated entities */}
  \STATE $e' \leftarrow e.max_{nei}$;
  \IF{$e' \in L$}
    \STATE $L \leftarrow L \cup \{e\}$;
  \ELSE 
    \STATE $\Omega \leftarrow avg(e.\Omega, e'.\Omega)$;
    \STATE $I \leftarrow I \cup \{<\Omega,(e,e')>\}$; // \textit{Hold the pair $(e,e')$ in $I$}
    \STATE $inter = \max(inter,\Omega)$;
  \ENDIF
\ELSE
  \WHILE{$e.\Omega < inter \ \& \ |L| < k-2$}
    \STATE $(e, e') = I.pop(inter)$; // \textit{Pop out the largest pair from $I$}
    $L \leftarrow L \cup \{e,e'\}$;
    \STATE $inter = \max(I.keys())$; // \textit{Update $inter$ of current $I$}
  \ENDWHILE
  \STATE $L \leftarrow L \cup \{e\}$;
\ENDIF
\STATE $i = i + 1$;
\ENDWHILE
\STATE \textbf{return} Landmark entity set $L$.
\end{algorithmic}
\end{algorithm}

\subsubsection{Complexity analysis}
Let the size of the candidate set be $|C|=n$. The complexity of ranking the candidates in descending order is $O(n\log(n))$.
When traversing the ranked candidate list, each traverse takes only $O(1)$ time. In the worst case, we need to scan the whole candidate list to generate the optimal landmark set $L^*$.
Therefore, the overall computational complexity of Algorithm \ref{alg:recalling} is $O(n\log(n)+n)$.


\section{Scalable GNN for Entity Alignment}\label{sect:gnn}
After acquiring the subgraphs of two KGs, we can start the embedding and alignment learning stage.
Instead of conducting message passing along the whole graph, our scalable GNN performs message passing within each subgraph and lets landmark entities bridge different subgraphs.
It largely reduces the computation complexity and the memory usage.
However, even if we try to compensate for the structure loss by recalling some important entities, the subgraph still loses some information compared with the complete graph.
We propose two methods to incorporate the interactions across subgraphs and further improve the performance.

\subsection{GNN Encoder}
For message passing, we use an efficient GNN model Dual-AMN \cite{mao2021boosting} as our encoder.
GNN-based entity alignment methods have also achieved cutting-edge performance thanks to their powerful structure learning ability \cite{EA_survey_AIOpen}.
In addition, there are also some other representation learning techniques like translational embedding models \cite{chen2017multilingual,sun2018bootstrapping} available for entity alignment.
But in our problem setting, 
GNNs are more extensible to represent entities when the neighborhood is incomplete.
For these reasons, we choose a GNN as the subgraph encoder to generate entity embeddings for alignment learning.
There are two layers in Dual-AMN, i.e., an inner graph layer and a cross graph layer.
The inner-graph layer uses a relational attention mechanism to capture structural information within a single KG.
The cross-graph layer connects entities with proxy nodes to capture alignment information.
The entity $e$ is represented as:
\begin{equation}
    \mathbf{e}^{\text{out}} = \texttt{Agg}_2\big(\texttt{Agg}_1(e, N_e), \mathcal{E}_{\text{proxy}}\big),
\end{equation}
where $\texttt{Agg}_1()$ denotes the neighborhood aggregator, 
while $\texttt{Agg}_2()$ combines the embeddings with proxy entities $\mathcal{E}_{\text{proxy}}$ to get the final outputs.
Interested readers can refer to its original paper \cite{mao2021boosting} for more details.
Please note that, our work does not seeks a powerful or novel GNN encoder.
We instead concentrate on the scalable and efficient GNN learning framework for entity alignment in large KGs.
During training, the GNN encoder is applied to each subgraph to obtain entity representations.
In contrast, the GNN in Dual-AMN is operated on the entire graph, which is not scalable to large KGs.

\subsection{Cross-subgraph Alignment Learning}
Alignment learning seeks to minimize the embedding distance of identical entities while separating the randomly sampled dissimilar entities using the output entity embeddings of the GNN encoder.
The seed entity pairs in training data are referred to as identical entities, whereas dissimilar entities are generated by negative sampling.
We also follow the loss function in Dual-AMN \cite{mao2021boosting} to achieve the alignment learning objective in each subgraph:
\begin{equation}\label{eq:align_loss}
\begin{split}
    \mathcal{L}_{\text{align}} = -\log \Big[&1 + \sum_{(e_1, e_2)\in \mathcal{A}_{s}} \sum_{(e_1, e_2')\in A'_{e_1}} \\ &\Big(\exp \big(\alpha (\beta + \pi(e_1, e_2) - \pi(e_1, e_2'))\big)\Big)\Big],
\end{split}
\end{equation}
where $A'_{e_1}$ represents the negative alignment pairs generated for entity $e_1$. 
$\alpha$ is a scale factor, and $\beta$ is the pre-defined margin for distinguishing between the identical entities and dissimilar entities.
$\pi()$ is a similar measure like Cosine.
A conventional method to generate dissimilar entities for an entity is random sampling \cite{sun2017cross}.
However, the randomly sampled dissimilar entities are usually ineffective for alignment learning, 
because they are typical really ``dissimilar'' and the model can easily distinguish between them. 
To mine hard negative alignment pairs for training,
The work \cite{sun2018bootstrapping} proposes the truncated negative sampling method that samples from the similar neighbors of the entity $e_1$.
In this way, the sampled negative examples also bear some similarity to $e_1$, increasing the difficulty of the model to distinguish between them. 
However, this method needs to compute the pairwise similarities between other entities and $e_1$, which is time-consuming, especially for large-scale KGs.
To resolve this issue, Dual-AMN treats all other entities in the current training batch as the dissimilar entities of $e_1$.
This method does not need sampling, and the similarity measure is inner product:
\begin{equation}\label{eq:sim}
\pi(e_1, e'_2) = \mathbf{e}_1^{\text{out}} \cdot \mathbf{e}_2^{\text{out}'},
\end{equation}
where $\mathbf{e}_1^{\text{out}}$ denotes the output embedding of the GNN encoder for $e_1$.
The embedding similarity between $e_1$ and all other in-batch entities can be quickly calculated by matrix multiplication.
Motivated by the effectiveness and efficiency of the method,
we use all other entities in the subgraph as the dissimilar entities of $e_1$ to generate negative alignment pairs, denoted as $A^{\text{intra\_neg}}_{e_1}$.
We can rewrite the $A'_{e_1}$ in Eq.~(\ref{eq:align_loss}) as $A^{\text{intra\_neg}}_{e_1}$ for alignment learning in the subgraph.

\subsubsection{Cross-subgraph negative alignment sampling and learning}
Alignment learning in a subgraph as shown in Eq.~(\ref{eq:align_loss}) does not consider the entities in other subgraphs.
In our alignment inference, we retrieve the counterpart of an entity from all the entities in the target KG, rather than only the entities in the subgraph.
Hence, we need to consider the entities from other subgraphs in alignment learning.
When training a subgraph as a mini-batch, we randomly sample $n$ entities from other subgraphs as the candidates for generating negative alignment pairs.
The candidate entities do not have output embeddings produced by the GNN encoder because their subgraphs are not available in this training batch.
Hence, we propose to use the input representations of entities and the loss for separating the sampled dissimilar entities is as follows:
\begin{equation}\label{eq:neg_align_loss}
    \mathcal{L}_{\text{cross}} = -\log \Big[1 + \sum_{e_1\in \mathcal{E}_i} \sum_{(e_1, e_2')\in A^{\text{cross\_neg}}_{e_1}} \exp \big(- \pi(e_1, e_2')\big)\Big],
\end{equation}
where $\mathcal{E}_i$ denotes the entity set of the $i$-th subgraph. $A^{\text{cross\_neg}}_{e_1}$ denotes the negative alignment pairs generated from cross-subgraph entities.
In this loss function, we only consider dissimilar entities.
The identical entities in training data are considered in Eq.~(\ref{eq:align_loss}). 
Embedding similarity is calculated using input representations.
\begin{equation}\label{eq:sim_sim_input}
\pi(e_1, e'_2) = \mathbf{e}_1^{\text{input}} \cdot \mathbf{e}_2^{\text{input}'},
\end{equation}
where $\mathbf{e}_1^{\text{input}}$ denotes the input representation of entity $e_1$.
In this way, we do not need to read the subgraphs of the randomly sampled cross-subgraph entities and feed them into the GNN encoder to get their output embeddings, for saving memory and training burden.

\subsubsection{Entity reconstruction from neighborhood}
An entity may appear in multiple subgraphs, and the partition process would cause the loss of some triples.
Hence, it is inevitable that the neighborhood of some entities in a subgraph is incomplete. 
We propose a self-supervised task, entity reconstruction, which enables the GNNs to reconstruct an entity representation with incomplete neighbors:
\begin{equation}\label{eq:reconstruct_loss}
    \mathcal{L}_{\text{reconstruct}} = \sum_{e \in \mathcal{E}_1\cup \mathcal{E}_2} \frac{1}{|{N}_{e}|} \sum_{e' \in {N}_e}  \| \mathbf{e} - \mathbf{e}' \|,
\end{equation}
where ${N}_{e}$ denotes the set of entity $e$'s neighbors in the subgraph.
The overall training loss is the combination of the alignment learning loss in Eq.~(\ref{eq:align_loss}), the loss negative alignment learning across subgraphs in Eq.~(\ref{eq:neg_align_loss}), and the entity reconstruction loss in Eq.~(\ref{eq:reconstruct_loss}).

\subsection{Graph-level Alignment Inference}
\subsubsection{Embedding space integration}
At the end, we have obtained the embedding of each entity within each subgraph.
We propose a graph-level embedding integration strategy to aggregate landmark entity embedding across subgraphs.
The final embedding representation of any $e \,in \,L$ that appears in multiple subgraphs is the average embedding of all subgraphs in which it appears.
This completes the representation of landmark entities because it already aggregates information from various subgraphs.
Also, during training, the landmark entities carry information from other subgraphs. This lets each subgraph get information from outside of itself, which gives message passing a more complete view of the information.

\subsubsection{Bidirectional $k$NN alignment search}
Conventional embedding-based methods, including LargeEA \cite{ge2021largeea} and LIME \cite{zeng2022entity}, assume that each entity in the source KG must have a counterpart in the target KG.
This ideal setting does not exist in the real world.
To find real entity alignment and reduce incorrect outputs,
we propose the bidirectional $k$NN alignment search that outputs two entities as an alignment pair only if they are the top-$k$ nearest neighbors of each other.
As the embedding space is very large, we use Faiss \cite{Faiss} for efficient approximated similarity search.
Compared to recent work \cite{DBP2} considering non-matchable entities, 
our alignment inference method does not require training and is scalable to large KGs.

\section{Experiment}

\subsection{Settings}
\subsubsection{Datasets}
We conduct experiments on datasets with different scales, which are small (15K), medium (100K) and large (1M), respectively. 
The detailed statistics are shown in Table~\ref{tb:dataset}.

\begin{table}[t]
\caption{Statistics of the datasets used in our experiments. 
15K and 100K datasets are from OpenEA \cite{sun2020benchmarking}.
The proposed \datasetname datasets are constructed based on DBpedia.}
\label{tb:dataset}
\renewcommand\arraystretch{1.2}
\resizebox{1.0\linewidth}{!}{
\begin{tabular}{|ll|l|l|l|}
\hline
\multicolumn{2}{|c|}{Datasets} & \# Entities & \# Relations & \# Triples \\ 
\hline
\multicolumn{1}{|l|}{\multirow{2}{*}{15K}}  & EN-FR & 15,000$-$15,000 & 267$-$210 & 47,334$-$40,864 \\
\multicolumn{1}{|l|}{} & EN-DE & 15,000$-$15,000 & 215$-$131 & 47,676$-$50,419 \\ \hline
\multicolumn{1}{|l|}{\multirow{2}{*}{100K}} & EN-FR & 100,000$-$100,000 & 400$-$300 & 309,607$-$258,285 \\
\multicolumn{1}{|l|}{} & EN-DE & 100,000$-$100,000 & 381$-$196 & 335,359$-$336,240 \\ \hline
\multicolumn{1}{|l|}{\multirow{2}{*}{1M}} & EN-FR & 1,211,270$-$1,176,869 & 595$-$400 & 8,150,656$-$4,268,719 \\
\multicolumn{1}{|l|}{} & EN-DE & 1,160,306$-$849,646 & 596$-$247 & 7,922,555$-$3,000,489\\ 
\hline
\end{tabular}}
\end{table}

\begin{table*}[!t]
\centering
\caption{Entity alignment results on our \datasetname.}
\label{exp:overall_1M}
\renewcommand\arraystretch{1.2}
\resizebox{0.7\textwidth}{!}{
\begin{tabular}{|l|l|ccc|ccc|cc|}
\hline
\multirow{2}{*}{Methods} & \multirow{2}{*}{Datasets} & \multicolumn{3}{c|}{Real setting}  & \multicolumn{3}{c|}{Ideal setting} & \multicolumn{2}{c|}{Efficiency} \\ 
\cline{3-10}
 &  & Precision & Recall & F1 & MRR & H@5 & H@1 & Time (s) & Memory (M) \\ 
\hline
\multirow{2}{*}{LargeEA} & EN-FR-1M & 0.357 & 0.073 & 0.121 & 0.224 & 0.251 & 0.127 & 796 & 17067 \\
 & EN-DE-1M & 0.428 & 0.109 & 0.174 & 0.237 & 0.282 & 0.125 & 687 & 17067 \\ 
\hline
\multirow{2}{*}{LIME} & EN-FR-1M & 0.364 & 0.080 & 0.131 & 0.232 & 0.298 & 0.166 & 2803 & 33963 \\
& EN-DE-1M & 0.448 & 0.111 & 0.178 & 0.275 & 0.338 & 0.213 & 2355 & 17067\\ 
\hline
\multirow{2}{*}{\modelname (CPS)} & EN-FR-1M & 0.389 & 0.132 & 0.195 & 0.232 & 0.276 & 0.138 & 1229 & 17067\\
 & EN-DE-1M & 0.480 & 0.131 & 0.206 & 0.258 & 0.311 & 0.157 & 988 & 17067 \\
 \hline
\multirow{2}{*}{\modelname (SBP)} & EN-FR-1M & 0.388 & 0.152 & 0.218 & 0.217 & 0.278 & 0.149 & 3299 & 33963\\
 & EN-DE-1M & 0.485 & 0.176 & 0.258 & 0.259 & 0.310 & 0.177 & 2684 & 17067 \\ 
\hline
\multirow{2}{*}{\modelname (w/o CLG)} & EN-FR-1M & 0.390 & 0.169 & 0.236 & 0.221 & 0.283 & 0.154 & 224 & 17067 \\
& EN-DE-1M & 0.485 & 0.188 & 0.271 & 0.263 & 0.326 & 0.193 & 179 & 17067\\
\hline
\multirow{2}{*}{\modelname} & EN-FR-1M & 0.393 & 0.183 & 0.250 & 0.245 & 0.314 & 0.178 & 2789 & 17067\\
& EN-DE-1M & 0.491 & 0.220 & 0.303 & 0.289 & 0.355 & 0.225 & 2018 & 17067 \\
\hline
\end{tabular}}
\end{table*}

\begin{list}{\labelitemi}{\leftmargin=1em}
    \item 15K and 100K datasets in OpenEA. 
    For small- and medium-scale datasets, we use the benchmark dataset (V1) in OpenEA \cite{sun2020benchmarking}. Unlike other datasets DBP15K \cite{sun2017cross} and DWY100K \cite{sun2018bootstrapping} that contain a large number of entities with high degrees, this benchmark follows the data distribution of real KGs. Following LargeEA \cite{ge2021largeea}, we select the two cross-lingual settings extracted from the multi-lingual DBpedia (English-to-French and English-to-German). It has two scales, i.e., 15K and 100K reference entity alignment pairs, respectively.
    
    \item Our \datasetname datasets. 
    The LargeEA dataset \cite{ge2021largeea} contains isolated entities.
    The LIME dataset \cite{zeng2022entity} has no non-matchable entities.
    Both of them are not the real scenarios for structure-based entity alignment.
    Hence, we built a large-scale dataset, namely \datasetname, by extracting the relation triplets and interlinguage links of multi-lingual DBpedia.
    We use the full triples of the English, French and German versions of DBpedia as data source.
    We removed the isolated entities and their links, as well as some long-tail entities without links.
    Our dataset has two settings: EN-FR-1M ($488,563$ alignment pairs), and EN-DE-1M ($418,436$ alignment pairs).
    More than half of their entities are non-matchable.
    
\end{list}

\subsubsection{Evaluation Setup.}
Following Dual-AMN \cite{mao2021boosting}, 
we use $30\%$ of the reference alignment as training data, $10\%$ as validation data, and the remaining as testing data. 
As our \datasetname contains non-matchable entities, 
we evaluate the entity alignment performance in two settings, i.e., the real and ideal settings. 
In the real setting, we do not know which entity has a counterpart in the other KG. 
We assess the performance by Precision, Recall and F1-score. 
In the ideal setting, the most common setting used by current entity alignment approaches, we only evaluate the performance of aligning matchable entities.
We use Hits@1, Hits@5 and mean reciprocal rank (MRR) as the metrics in this setting. 
For all evaluation metrics, higher scores indicate better performance. 
We also report the training time in seconds and the maximum GPU memory cost in MB to evaluate the efficiency and scalability of our approach.

\subsubsection{Implementation Details}
The corresponding hyper-parameter values of the GNN encoder follow the settings on Dual-AMN \cite{mao2021boosting}. 
$\eta=0.001$ in Eq.~(\ref{equ: importance}). 
We do not label the importance of all entities. 
When the value of $\Phi(e)$ is smaller than $0.49$, 
we stop the labeling process. 
$\lambda=0.01$ in Eq.~(\ref{benefit}). 
The size of landmark set $L$ depends on the maximum subgraph size, we set it as $7500$, $25000$, and $90000$ for 15K, 100K and DBpedia1M dataset respectively. 
For our method, we set the partition number for the three datasets as 5, 10, and 40, respectively. The $k$ in bidirectional $kNN$ alignment search is set as $5$. We release the source code at Github\footnote{\url{https://github.com/JadeXIN/LargeGNN}}.


\subsubsection{Baselines}
As most embedding-based approaches fail to be implemented on large-scale KGs, 
we only compare \modelname with two scalable approaches, LargeEA \cite{ge2021largeea} and LIME \cite{zeng2022entity}.
Both of them follow the ``blocking-then-learning'' framework and do entity alignment in each subgraph independently. 
For fair comparison, we implement our GNN model on their framework to guarantee the consistency.
To validate the effectiveness of our subgraph generation strategy, we also compare with the partitioning methods CPS in LargeEA \cite{ge2021largeea} and SBP in LIME \cite{zeng2022entity}.
As we evaluate our performance on graph level, 
we implement them in our approach by \modelname (CPS) and \modelname (SBP).
As our approach is solely based on structure information and supervised setting, 
we evaluate these methods in the same setting.
We do not consider the name channel from LargeEA, and the semi-supervised setting I-SBP from LIME. The use of textual information and semi-supervised learning are left for future work.
The ablation study variants are as follows:
\modelname denotes our full approach, which means our GNN module with the Centrality Subgraph Generation (CSG) strategy;
\modelname(w/o CLG) means removing the connectivity-aware landmark generation (CLG) process, which only conducts the seed-targeted KG merging and partitioning;
\modelname(w/o CNS) and \modelname(w/o ER) are used to evaluate our cross-subgraph negative sampling (CNS) and entity reconstruction (ER). 

\subsection{Results and Analyses}

\begin{table}[!t]
\caption{Entity alignment results on OpenEA 100K datasets.}\label{exp:overall_100k}
\renewcommand\arraystretch{1.3}
\resizebox{1.0\linewidth}{!}{
\begin{tabular}{|l|l|cccc|ll|}
\hline
\multirow{2}{*}{Methods} & \multirow{2}{*}{Datasets} & \multicolumn{4}{|c|}{Ideal setting} & \multicolumn{2}{|c|}{Efficiency} \\ 
\cline{3-8}
 & & H@1 & Train & Test & S-size & Time & Mem. \\ \hline
\multirow{2}{*}{LargeEA} & EN-FR-100K & 0.300 & 0.747 & 0.574 & 20000 & 15.2 & 4811 \\
\multirow{2}{*}{} & EN-DE-100K & 0.255 & 0.751 & 0.583 & 20000 & 18.5 & 4795 \\ \hline
\multirow{2}{*}{LIME} & EN-FR-100K & 0.397 & 0.962 & 0.875 & 35231 & 43.2 & 4779 \\
\multirow{2}{*}{} & EN-DE-100K & 0.460 & 0.954 & 0.860 & 35432 & 45.3 & 4779 \\ \hline
\multirow{2}{*}{\modelname (CPS)} & EN-FR-100K & 0.337 & 0.779 & 0.635 & 20000 & 21.6 & 4779 \\
\multirow{2}{*}{} & EN-DE-100K & 0.325 & 0.772 & 0.641 & 20000 & 24.8 & 4779 \\ \hline
\multirow{2}{*}{\modelname (SBP)} & EN-FR-100K & 0.384 & 0.964 & 0.888 & 35231 & 35.1 & 4779 \\
\multirow{2}{*}{} & EN-DE-100K & 0.473 & 0.971 & 0.884 & 35432 & 50.3 & 4799 \\ \hline
\multirow{2}{*}{\modelname(w/o CLG)} & EN-FR-100K & 0.403 & 1.000 & 0.861 & 20000 & 14.9 & 4779 \\
\multirow{2}{*}{} & EN-DE-100K & 0.491 & 1.000 & 0.859 & 20000 & 14.2 & 4779 \\\hline
\multirow{2}{*}{\modelname} & EN-FR-100K & 0.423 & 1.000 & 0.893 & 25000 & 24.3 & 4779 \\
\multirow{2}{*}{} & EN-DE-100K & 0.544 & 1.000 & 0.883 & 25000 & 43.2 & 4779 \\
\hline
\end{tabular}}
\end{table}

\subsubsection{Results on large-scale datasets}

Table \ref{exp:overall_1M} presents the entity alignment results in both real and ideal settings on our proposed DBpedia1M, along with the corresponding partition time and memory usage. 
\modelname achieves the best performance. 
It outperforms the second best baseline \modelname (SBP) by $3\%$ on EN-FR-1M and $4\%$ on EN-DE-1M by F1 measure, as well as $3\%$ on EN-FR-1M and $5\%$ on EN-DE-1M by Hits@1. 
The main reason of our promising performance on such a large dataset is that both our effective subgraph generation strategy and the cross-subgraph training strategies can reduce the structure loss caused by graph partitioning. 
To compare with the ``blocking-then-learning'' framework, we can observe the performance of \modelname (CPS) and LargeEA, or \modelname (SBP) and LIME.
They have the same graph partition strategy. 
CPS can achieve a large improvement in our approach, because our graph-level alignment inference can largely reduce the influence of the lost alignment pairs from the CPS partitioning strategy.
However, the improvement of SBP in our approach is not obvious.
This is because the subgraphs generated by SBP have much overlap.
The large ratio of overlapping can bring a large improvement in its own learning framework as it conducts alignment inference only at subgraph level.
But the overlapping may have a bad influence on our approach because the entities that carry incomplete training signals may have a bad effect on our subgraph training process.
Overall, SBP performs better than CPS as the baselines that utilize SBP achieve better performance than those that use CPS. This is owing to the bidirectional partitioning process, which can recall a large number of seed entity pairs.
Therefore, the test pair recall is increased correspondingly. But SBP consumes more than double the time than CPS, as it conducts twice the CPS partition process and the union of subgraph results of two partition processes, which also leads to a larger subgraph size.
The larger subgraph size may result in larger memory consumption, such as on EN-FR-1M.

As for scalability, we can see that CPS is the fastest. 
The reason why CPS and SBP in \modelname cost longer time than in their original approaches is that their subgraph pair formalization process may discard some subgraphs, 
as it pairs subgraphs in different KGs by matching the one with the largest seed alignment entities. 
Therefore, we utilize a stable matching process to guarantee that no entities are missed in our implementation. 
SBP has more than twice the time cost of LargeEA, but it leads to better entity alignment results.
The extra time cost is also acceptable.
It totally costs less than an hour. 
The partitioning time costs of \modelname and SBP are similar, and ours achieves higher alignment performance.

From the last two lines of table \ref{exp:overall_1M}, we can see our CLG process can improve the EA result by $1.4\%$ on EN-FR-1M and $3.2\%$ on EN-DE-1M. 
This is because our CLG process can select those entities with important learning signals and enhance the inter-communication among subgraphs. However, this process costs extra time. 
We can see that, without CLG, the process of merging seed entities and partitioning only takes around 200 seconds. 
The large time cost is mainly comes from the entity importance labeling process in Section \ref{sec: importance}, which needs to label the importance compatibility. 

\begin{table}[t]
\caption{Entity alignment results on OpenEA 15K datasets.}\label{exp:overall_15k}
\renewcommand\arraystretch{1.3}
\resizebox{1.0\linewidth}{!}{
\begin{tabular}{|l|l|cccc|ll|}
\hline
\multirow{2}{*}{Methods} & \multirow{2}{*}{Datasets} & \multicolumn{4}{|c|}{Ideal setting} & \multicolumn{2}{|c|}{Efficiency} \\ 
\cline{3-8}
 & & H@1 & Train & Test & S-size & Time & Mem. \\ \hline
\multirow{2}{*}{LargeEA} & EN-FR-15K & 0.313 & 0.764 & 0.620 & 6000 & 1.4 & 2239 \\
\multirow{2}{*}{} & EN-DE-15K & 0.552 & 0.922 & 0.847 & 6000 & 1.8 & 2239 \\ \hline
\multirow{2}{*}{LIME} & EN-FR-15K & 0.545 & 0.962 & 0.966 & 96357 & 3.1 & 2253 \\
\multirow{2}{*}{} & EN-DE-15K & 0.656 & 0.993 & 0.963 & 94234 & 3.8 & 2253 \\ \hline
\multirow{2}{*}{\modelname (CPS)} & EN-FR-15K & 0.489 & 0.775 & 0.646 & 6000 & 1.4 & 2247 \\
\multirow{2}{*}{} & EN-DE-15K & 0.606 & 0.935 & 0.858 & 6000 & 1.7 & 2141 \\ \hline
\multirow{2}{*}{\modelname (SBP)} & EN-FR-15K & 0.556 & 0.970 & 0.904 & 96357 & 2.5 & 2141 \\
\multirow{2}{*}{} & EN-DE-15K & 0.707 & 0.982 & 0.965 & 94234 & 3.5 & 2141 \\ \hline
\multirow{2}{*}{\modelname (w/o CLG)} & EN-FR-15K & 0.563 & 1.000 & 0.919 & 6000 & 0.8 & 2141 \\
\multirow{2}{*}{} & EN-DE-15K & 0.710 & 1.000 & 0.948 & 6000 & 1.0 & 2141 \\\hline
\multirow{2}{*}{\modelname} & EN-FR-15K & 0.576 & 1.000 & 0.957 &  7500 & 2.8 & 2141 \\
\multirow{2}{*}{} & EN-DE-15K & 0.746 & 1.000 & 0.962 & 7500 & 3.1 & 2141 \\
\hline
\end{tabular}}
\end{table}

\subsubsection{Results on small and medium datasets}
As there are no non-matchable entities in small and medium datasets, we only evaluate the performance in the ideal setting. 
We also test the ratio of preserved alignment pairs of training data and test data, shown as ``Train'' and ``Test'', respectively, and record the average size of subgraphs, denoted as ``S-size''. 
The overall results are shown on Tables \ref{exp:overall_100k} and \ref{exp:overall_15k}. 
We can see the results of large dataset is much worse than that of small and medium datasets. Besides large KG has more complex structure, the large amount of non-matchable entities also bring big challenges for entity alignment.
\modelname achieves the best performance on these datasets. 
It outperforms the second best baseline by approximately $2\%$ to $7\%$. 
\datasetname is more challenging than these datasets due to its large candidate space and large number of non-matchable entities.

In terms of memory usage, all the baselines have similar memory costs, even their subgraph size varies. 
This is because memory consumption is relevant to many factors, including the GPU cache. 
When the subgraph size is within a specific range, the memory usage remains unchanged. 
We have a detailed discussion in the next section.
As for efficiency, the time cost of these baselines on small and medium datasets follow a similar trend to that of large datasets
where our method without CLG is still the fastest, and CPS is the second fastest. 
SBP costs more than twice as much as CPS. 
Ours consumes more time than CPS but less than SBP.

Finally, the trend of effectiveness is also similar to the result of large datasets. 
The partition method of LIME outperforms LargeEA. 
However, the improvement of entity alignment is at the cost of more than doubled time and possibly memory increase, even though it has the same memory consumption in this case. 
Our method merges two KGs into a single graph, which avoids the double partitioning process, and guarantees 100\% seed entity pairs.
It leads to increased test pair recall without too much increased subgraph sizes. 
Comparing our method with the one removing the CLG process, 
\modelname also has $1\%$ to $5\%$ improvement on Hits@1, which again validates the effectiveness of CLG. 
We can see that only merging seed entity pairs can lead to higher results than SBP, even its alignment pair recall score is lower than that of SBP. 
Compared to improving the number of preserved alignment pairs in each subgraph, recalling entities that carry more structure information is more important. 

\subsubsection{Scalability discussion}
Overview, our approach with or without CLG has excellent scalability. 
First, by simply merging the seed alignment pairs, it only costs less than an hour on a million-scale dataset, and achieves better results than both CPS and SBP, even though SBP has larger subgraph size. 
Second, the CLG process can further improve entity alignment performance by recalling landmark entities, at the cost of a bit longer time, but still acceptable, compared with the slowest partition SBP.

\subsection{Ablation Study}

\begin{figure}
\centering
  \includegraphics[width=0.42\linewidth]{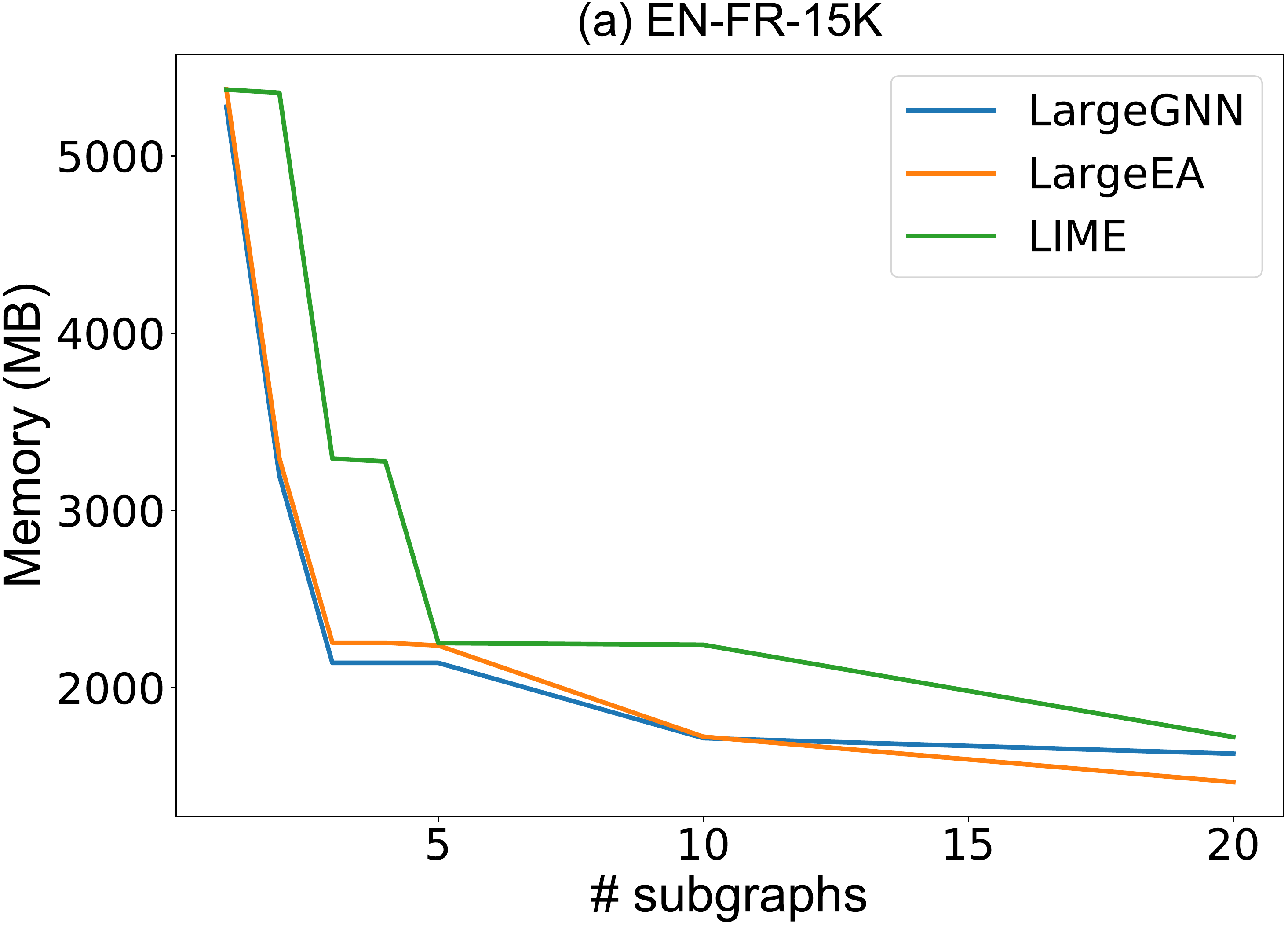}
  \hspace{2pt}
  \includegraphics[width=0.48\linewidth]{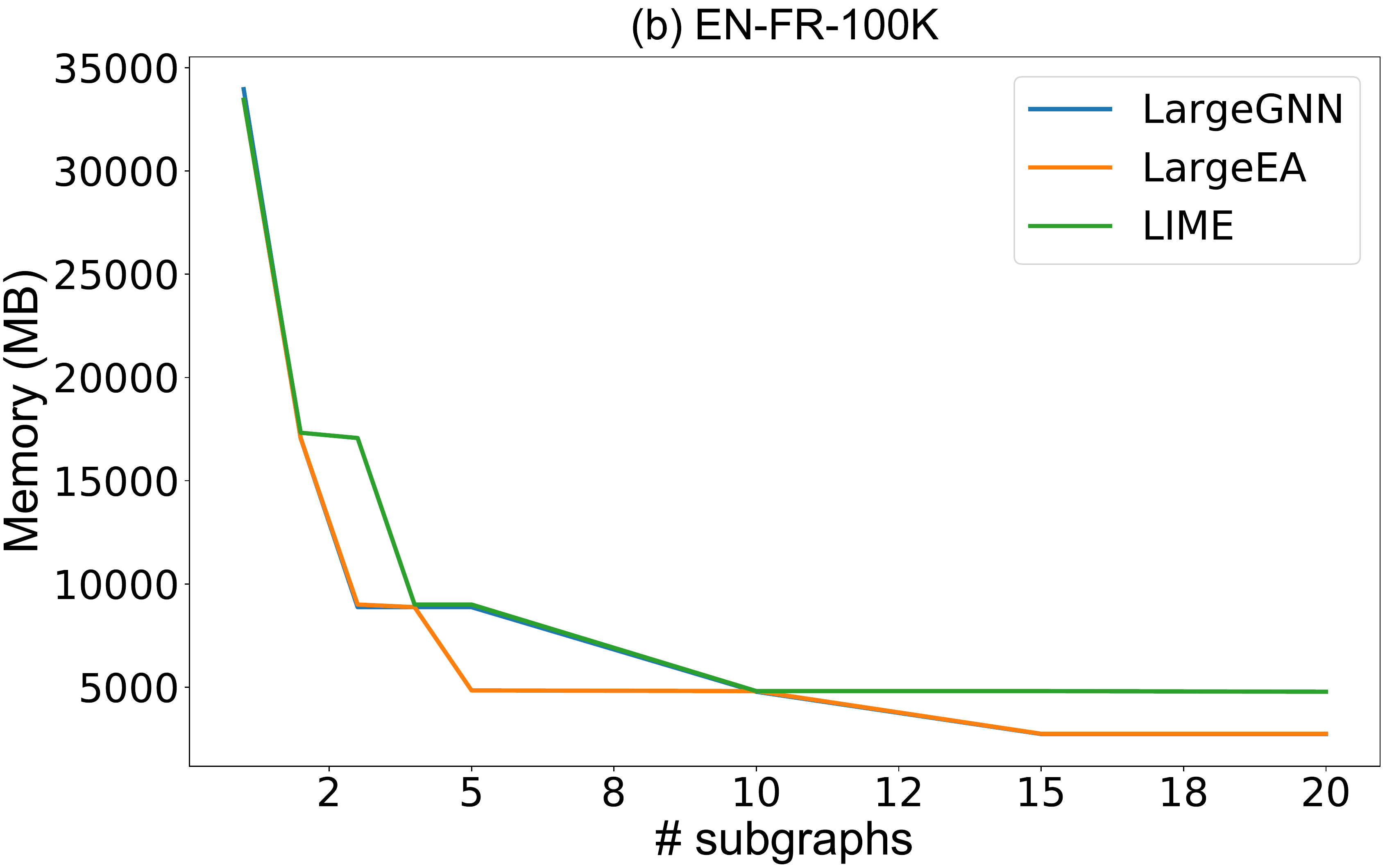}
\caption{\label{fig:mem}Memory usage w.r.t. different \# subgraphs.}
\end{figure}

\begin{figure}
\centering
  \includegraphics[width=0.45\linewidth]{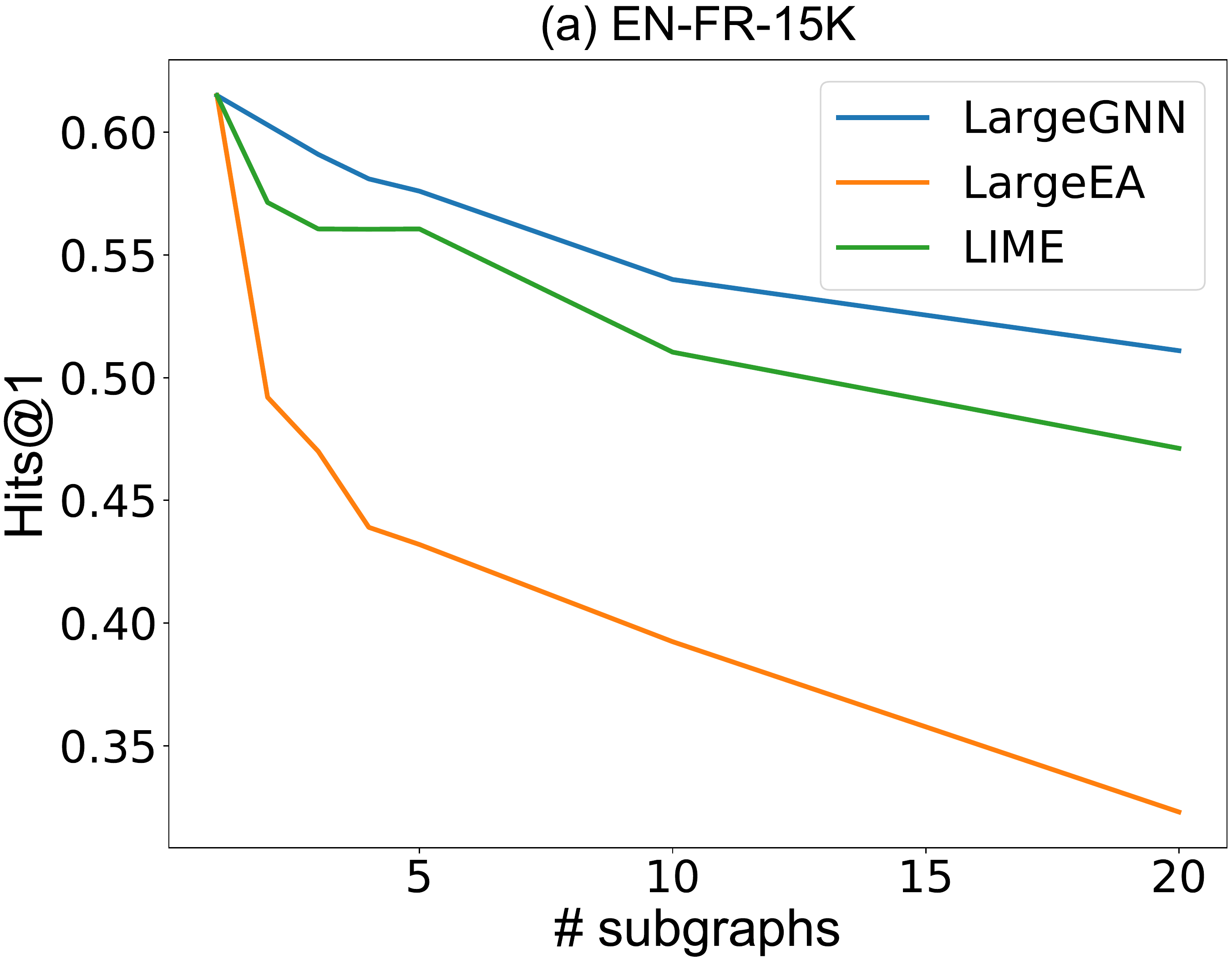}
  \hspace{2pt}
  \includegraphics[width=0.45\linewidth]{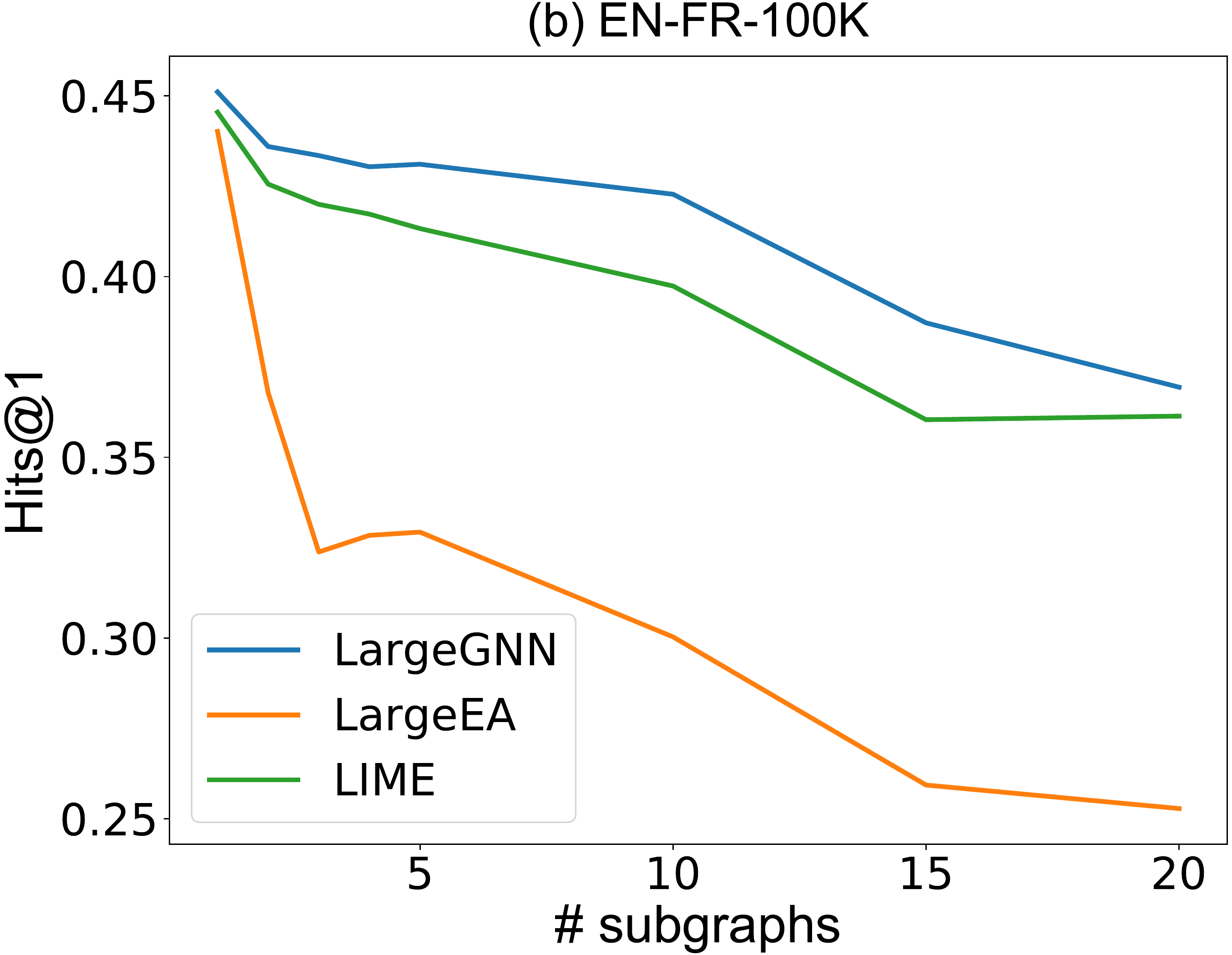}
\caption{\label{fig:acc}Hits@1 w.r.t. different \# subgraphs.}
\end{figure}

\begin{figure}
    \centering
      \includegraphics[width=0.46\linewidth]{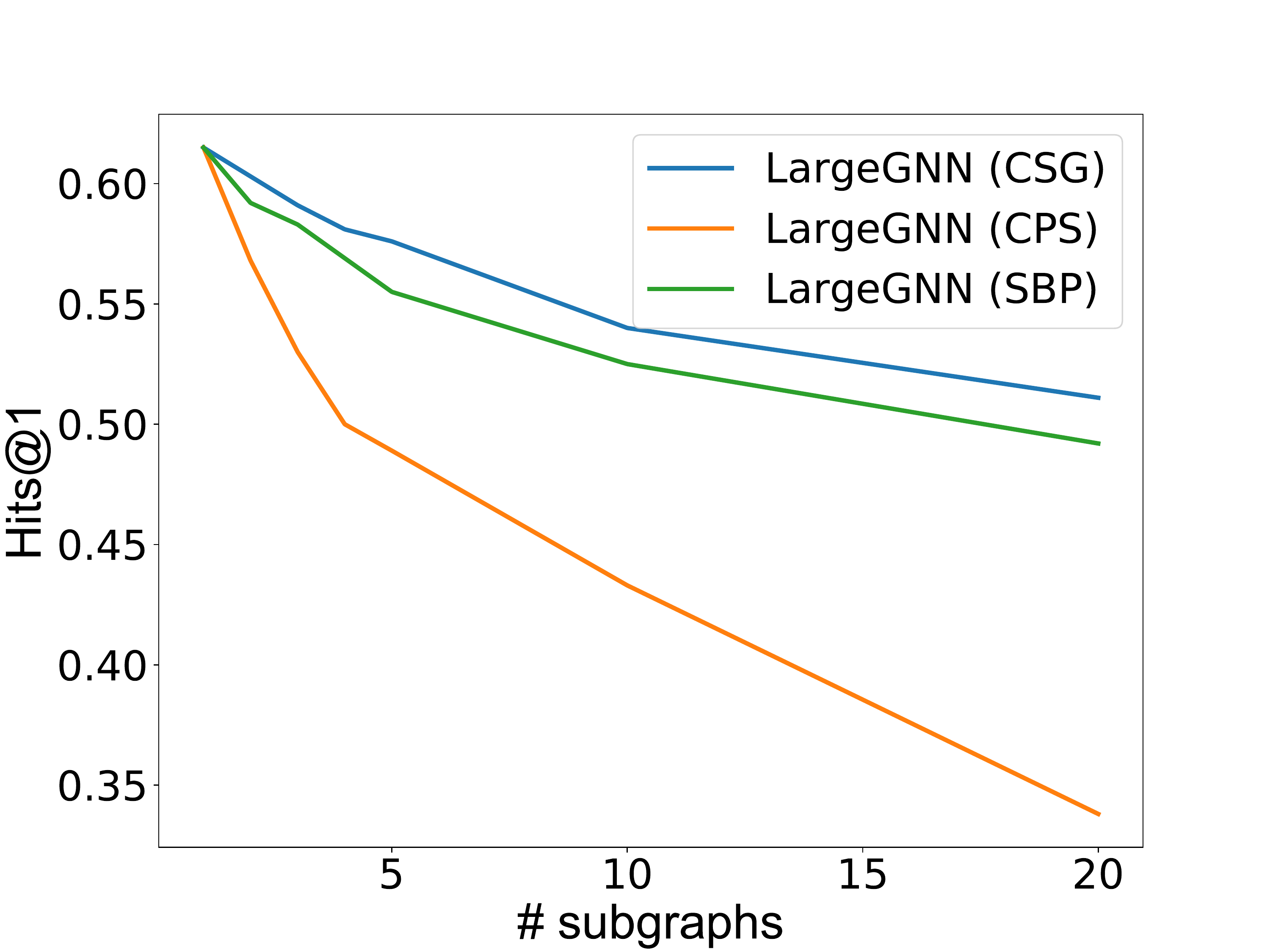}
  \hspace{2pt}
  \includegraphics[width=0.45\linewidth]{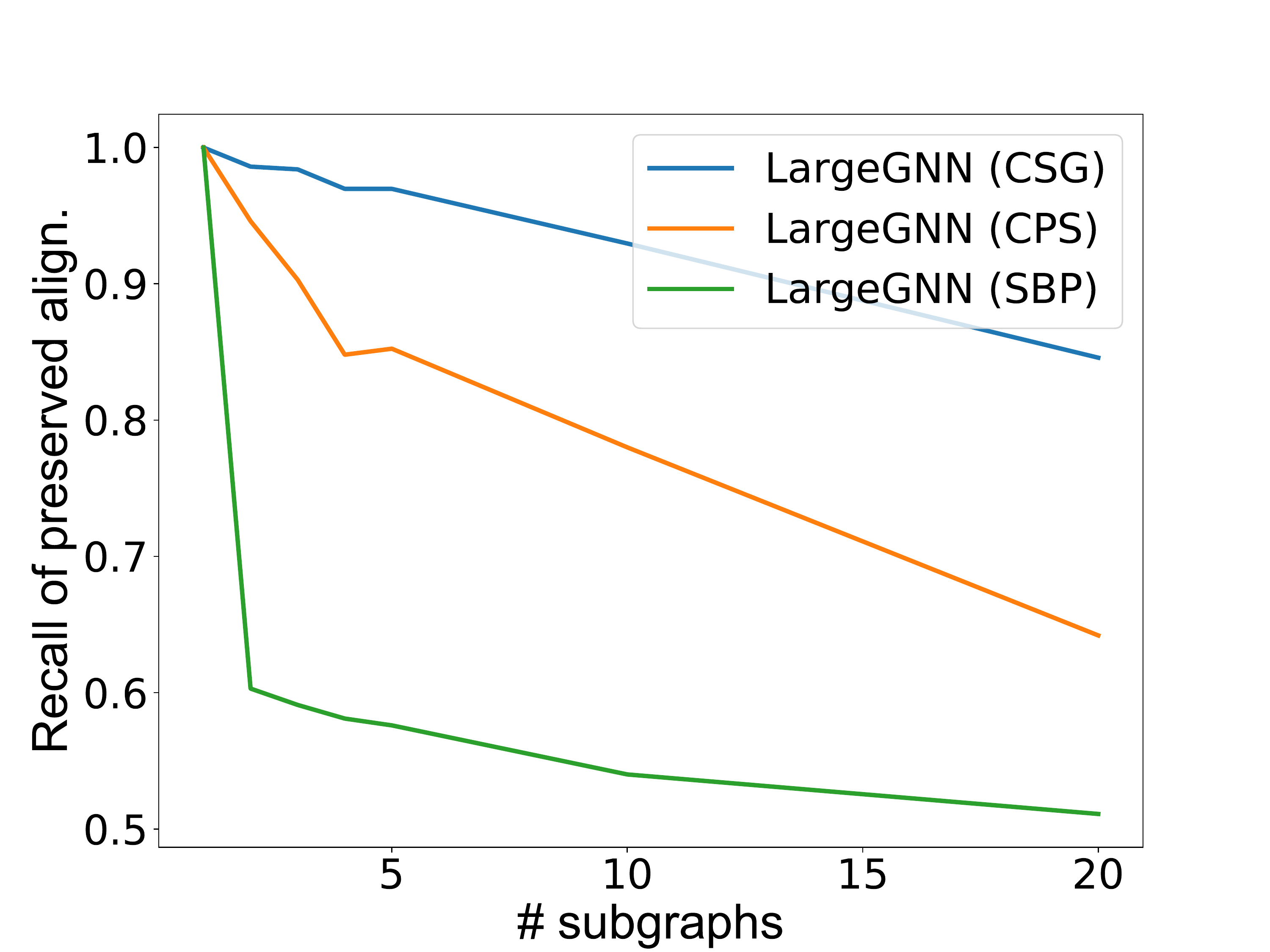}
    \caption{\label{fig:3partition}Hits@1\,and\,recall of preserved align. on EN-FR-15K.}
\end{figure}

\subsubsection{Comparison of three scalable methods}
We test the influence of different number of subgraphs on our method and the two baselines, LargeEA and LIME. 
Figure \ref{fig:mem} shows the memory usage as the number of subgraphs increases. 
The trend of the three are similar.
When the subgraph number is within the range of 3 to 5, 
the memory usage almost remains unchanged. 
That is why LIME generates larger subgraph sizes but has the same memory usage as other baselines, since the cluster size does not exceed the range that needs more memory consumption. 
LIME sometimes occupies higher memory, because it takes the union of the subgraph results of two partitions, which results in a larger subgraph size.

Figure \ref{fig:acc} gives the Hits@1 of the three methods w.r.t. different numbers of subgraphs. 
As the number of subgraphs increases, 
the performance of all of them decreases. 
LargeEA has the highest decline, because it cannot protect the seed alignment properly. 
LIME has a smaller decrease because the bidirectional partition can largely increase the number of protected entity pairs. 
However, our method remains the best performance along with any number of subgraphs, 
and our subgraph sizes are smaller than LIME, which means we will not consume more memory.

\subsubsection{Comparison of different partition strategies}
Figure \ref{fig:3partition} compares the three partition strategies: our CSG, as well as CPS and SBP from LargeEA and LIME, respectively.
As the number of subgraphs increases, the recall of preserved alignment pairs by all partition strategies decreases, and CPS decreases the most. 
The higher recall of CSG benefits from the seed merging strategy, which can largely protect alignment pairs. Correspondingly, the Hits@1 performance drops as the recall drops, where largeEA also drops the largest amount. 
Even if SBP increases the recall by increasing the subgraph size, it still has lower performance than CSG. 
In our graph-level entity alignment setting, protecting alignment pairs is not the only standard.
Instead, how to recall more entities containing rich alignment information is more important.

\subsubsection{Effect of cross-subgraph learning}
Table \ref{tab:cross-subgraph} evaluates the effect of our two cross-subgraph training strategies: cross-subgraph negative sampling (CNS) and entity reconstruction (ER).
Both strategies can improve the performance. 
The ER strategy contributes more than CNS. 
The two strategies together can bring approximately $2\%$ improvement, where ER can improve by about $1.3\%$ and CNS about $0.7\%$. 
The reason of the improvement is that these two cross subgraph training strategies can bring training signals from other subgraphs, which further enhances the connection among subgraphs and attenuates the structure loss caused by graph partitioning.

\begin{table}[!t]
\centering
\caption{Hits@1 w.r.t. different GNN subgraph strategies.}
\resizebox{0.999\linewidth}{!}{
\renewcommand\arraystretch{1.2}
\begin{tabular}{|l|c|c|c|c|} 
\hline
Methods & EN-FR-15K & EN-DE-15K & EN-FR-100K & EN-DE-100K \\
\hline
\modelname & 0.576 & 0.746 & 0.423 & 0.544\\ 
\,\, w/o CNS & 0.569 & 0.739 & 0.419 & 0.539\\
\,\, w/o ER & 0.565 & 0.735 & 0.408 & 0.531\\
\,\, w/o CNS \& ER & 0.559 & 0.728 & 0.405 & 0.527\\
\hline
\end{tabular}}
\label{tab:cross-subgraph}
\end{table}

\section{Conclusion and Future Work}
In this work, we focus on the scalability issue of large-scale entity alignment. 
Our proposed approach \modelname can be effectively implemented on a million-scale KG. 
To mitigate the structure loss caused by graph partitioning, we complement it from three stages. 
First, at subgraph generation stage, we design a centrality-based subgraph generation algorithm, which recalls some influential entities to mitigate the structure loss. 
Then, we enhance the connection among subgraphs by two cross-subgraph training strategies. 
Final, we conduct the alignment inference on the whole graph. Extensive experiments validate the effectiveness and efficiency of \modelname. 
In future work, we will further improve the computation efficiency, and extend our approach by utilizing extra information.

\begin{acks}
This work was partially supported by the Australian Research Council under Grant No. DE210100160 and DP200103650.
\end{acks}

\bibliographystyle{ACM-Reference-Format}
\bibliography{main}
\end{document}